\documentclass[sigconf]{acmart}

\usepackage{epsfig}
\usepackage{graphicx}
\usepackage{amsmath}
\usepackage{helvet}  
\usepackage{courier}  
\usepackage{url}  
\usepackage{xcolor}
\usepackage{multirow}
\usepackage{subcaption}
\usepackage{array}

\usepackage{times}

\usepackage{balance}

\newcolumntype{P}[1]{>{\centering\arraybackslash}p{#1}}

\def\BibTeX{{\rm B\kern-.05em{\sc i\kern-.025em b}\kern-.08emT\kern-.1667em\lower.7ex\hbox{E}\kern-.125emX}}

\copyrightyear{2021}
\acmYear{2021}
\setcopyright{acmlicensed}
\acmConference[ICMR '21]{Proceedings of the 2021 International Conference on Multimedia Retrieval}{August 21--24, 2021}{Taipei, Taiwan}
\acmBooktitle{Proceedings of the 2021 International Conference on Multimedia Retrieval (ICMR '21), August 21--24, 2021, Taipei, Taiwan}
\acmPrice{15.00}
\acmDOI{10.1145/nnnnnnn.nnnnnnn}
\acmISBN{978-x-xxxx-xxxx-x/YY/MM}    
    
\begin{document}

\fancyhead{}
%
\title{Contextualized Keyword Representations for Multi-modal Retinal Image Captioning}

%

\author{Jia-Hong Huang}
\affiliation{%
  \institution{University of Amsterdam, Netherlands}}
\email{j.huang@uva.nl}

\author{Ting-Wei Wu}
\affiliation{%
  \institution{Georgia Institute of Technology, USA}}
\email{waynewu@gatech.edu}


 





\author{Marcel Worring}
\affiliation{\institution{University of Amsterdam, Netherlands}}
\email{m.worring@uva.nl}

%

%
\begin{abstract}

Medical image captioning automatically generates a medical description to describe the content of a given medical image. A traditional medical image captioning model creates a medical description only based on a single medical image input. Hence, an abstract medical description or concept is hard to be generated based on the traditional approach. Such a method limits the effectiveness of medical image captioning. Multi-modal medical image captioning is one of the approaches utilized to address this problem. In multi-modal medical image captioning, textual input, e.g., expert-defined keywords, is considered as one of the main drivers of medical description generation. Thus, encoding the textual input and the medical image effectively are both important for the task of multi-modal medical image captioning. In this work, a new end-to-end deep multi-modal medical image captioning model is proposed. Contextualized keyword representations, textual feature reinforcement, and masked self-attention are used to develop the proposed approach. Based on the evaluation of the existing multi-modal medical image captioning dataset, experimental results show that the proposed model is effective with the increase of +$53.2$\% in BLEU-avg and +$18.6$\% in CIDEr, compared with the state-of-the-art method.
\textbf{\href{https://github.com/Jhhuangkay/Contextualized-Keyword-Representations-for-Multi-modal-Retinal-Image-Captioning}{\scriptsize{https://github.com/Jhhuangkay/Contextualized-Keyword-Representations-for-Multi-modal-Retinal-Image-Captioning}}.}
\end{abstract}

%
%


%

%


%
\maketitle

\begin{figure}[ht]
\begin{center}
\includegraphics[width=1.0\linewidth]{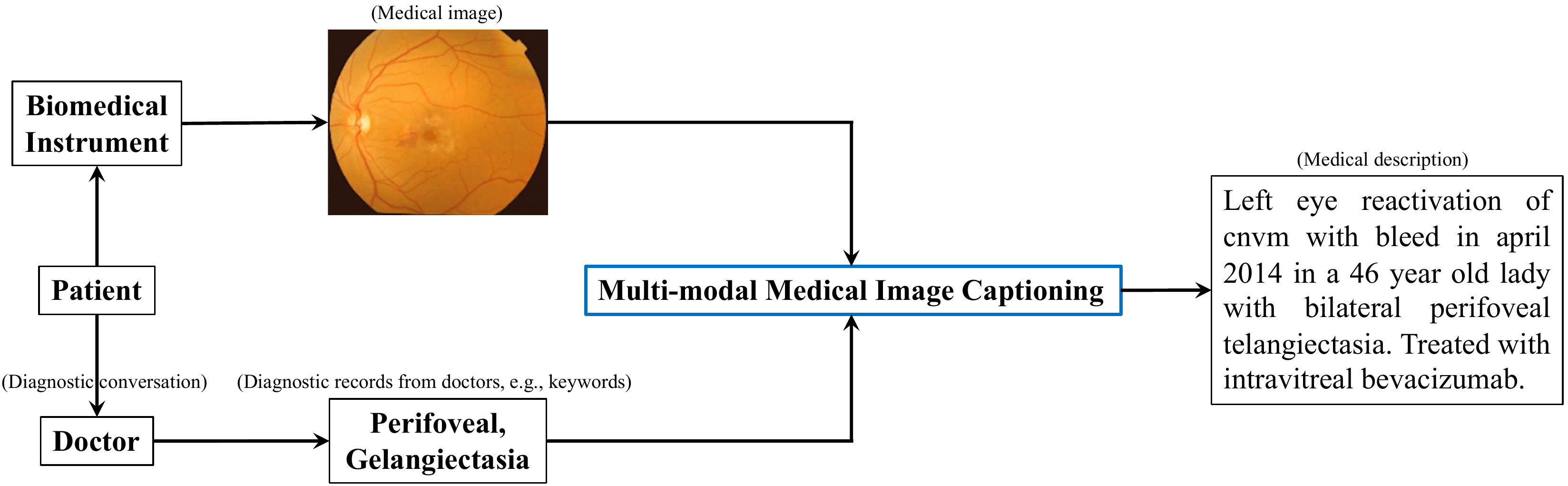}
\end{center}
\vspace{-0.5cm}
   \caption{Multi-modal medical image captioning. 
   A multi-modal medical image captioning algorithm takes a medical image, e.g., a retinal image, and text-based diagnostic records, e.g., a set of keywords, as inputs to generate a medical description. The effectiveness of traditional medical image captioning models is improved by the additional input keywords.
   }
\vspace{-0.3cm}
\label{fig:figure1}
\end{figure}

\section{Introduction} 
Medical image captioning automatically generates a medical report/description that describes the content of a given medical image  \cite{laserson2018textray,jing2018automatic,li2018hybrid,huang2021deepOpht}. However, traditional medical image captioning methods, e.g., \cite{laserson2018textray,li2018hybrid,jing2018automatic}, generate a medical description based on given image information only. It is hard to derive abstract medical descriptions/concepts \cite{laserson2018textray,jing2018automatic} based on the traditional approaches. Hence, they reduce the effectiveness of medical image captioning.

Multi-modal medical image captioning has been proposed as an approach to make conventional medical image captioning more effective \cite{huang2021deepOpht}. The main idea of multi-modal medical image captioning is to generate descriptions for a given medical image based on the additional text-based information provided by the doctor, e.g., using keywords to help the medical description generation, visualized in Figure \ref{fig:figure1}. According to \cite{huang2021deepOpht}, keywords commonly exist and they are from the doctors' textual diagnosis records in the early diagnosis process. Traditional medical image captioning only has one input modality, i.e., image, while an efficient choice for multi-modal medical image captioning is a set of keywords, in addition to medical image \cite{huang2021deepOpht}. Since the set of keywords is considered as one of the main inputs of multi-modal medical image captioning \cite{huang2021deepOpht}, effectively embedding the keywords is important. In \cite{huang2021deepOpht}, the Bag of Words (BoW) approach is used to encode the keywords input for multi-modal medical image captioning. BoW has been used with great success on many natural language processing (NLP) tasks, e.g., language modeling and document classification, but the authors of \cite{scott1998text,soumya2014text} point out that BoW suffers from the following. First, from the time and space complexity perspective, sparse representations are harder to model by BoW, so the vocabulary requires careful design. If the information in a large representation space is sparse, BoW models are not effective enough. Second, the semantic meaning likely cannot be captured effectively. 

In this work, a new approach is proposed that tackles the aforementioned issue to improve the performance of a multi-modal medical image captioning model. According to \cite{ethayarajh2019contextual}, the commonly used method of static word embeddings, e.g., global vectors for word representation (GloVe) \cite{pennington2014glove} or skip-gram with negative sampling (SGNS) \cite{mikolov2013distributed}, is a better choice to encode the textual input than BoW. However, a limitation with static word embeddings is that all senses of a polysemous word must share a single vector because GloVe and SGNS generate a single representation for each word \cite{ethayarajh2019contextual}. As stated in \cite{ethayarajh2019contextual} the approach of contextualized word representations, e.g., Generative Pretrained Transformer-2 (GPT-2), is more effective than static word embeddings. For encoding textual input, the proposed method exploits the contextualized word representations approach, i.e., GPT-2, to more effectively encode input keywords. For visual feature extraction, a pre-trained convolutional neural network (CNN), e.g., VGG16 or VGG19 pre-trained on ImageNet \cite{simonyan2014very,russakovsky2015imagenet}, is applied to effectively encode input medical image. Note that experiments conducted in this paper are based on the existing multi-modal retinal image captioning dataset which is proposed by \cite{huang2021deepOpht}. Retinal images from \cite{huang2021deepOpht}'s dataset are mostly colorful. Hence, using a CNN pre-trained on ImageNet is helpful in the experiments in terms of low-level features, e.g., color \cite{huang2021deepOpht}. According to the experimental results, the proposed multi-modal medical image captioning method generates a more accurate and meaningful description for a given retinal image than baselines.

\vspace{+3pt}
\noindent\textbf{Contributions.}
\begin{itemize}

    \item A new end-to-end deep model for multi-modal medical image captioning is proposed, based on the contextualized keyword representation, textual feature reinforcement module, and masked self-attention mechanism.

    \item The proposed method is thoroughly validated through experiments on the existing multi-modal retinal image captioning dataset.
    The experimental results show that the proposed model is more effective than the method based on static word embedding or BoW. The model performance is increased in terms of BLEU and CIDEr.

\end{itemize}

\noindent
The rest of this paper is organized as follows: In Section 2, the related work is reviewed. Then, the proposed approach is introduced in Section 3. Finally, an evaluation of the effectiveness of the proposed model is conducted in Section 4, followed by an analysis of the experimental results.

\section{Related Work}
In this section, the related works, i.e., image captioning, multi-modal medical image captioning, methods of word embeddings, and retinal image dataset are reviewed.

\subsection{Image Captioning}
The goal of conventional image captioning is to automatically generate a text-based description for a given natural image \cite{vinyals2015show,fang2015captions,karpathy2015deep}. In \cite{vinyals2015show}, an image captioning model with an encoder-decoder architecture was introduced. A CNN model was considered as an encoder to extract image features. A recurrent neural network (RNN) was considered as a decoder to produce a description for a given natural image, based on the extracted image feature. In \cite{fang2015captions}, a language model was exploited to combine a set of possible words, related to several parts of an input image, and generate a description of the image. The authors of \cite{karpathy2015deep} have introduced an approach that embeds language and visual information into a common space. In \cite{hendricks2016generating}, the authors have proposed a method that focuses on discriminating properties of the visible object. The proposed approached jointly predicts a class label and is used to explain why the predicted label is proper for a given input image. Based on reinforcement learning and sampling and through a loss function, their proposed model learned to generate captions for the given image. According to \cite{liu2017improved}, existing image captioning models are trained via maximum likelihood estimation. However, a limitation is that the log-likelihood score of some descriptions cannot well correlated with human assessments of quality. In \cite{gao2019deliberate}, a deliberate residual attention image captioning model was proposed. In the proposed model, the layer of first-pass residual-based attention was used to generate hidden states and visual attention. A preliminary image description was then created. The layer of second-pass deliberate residual-based attention was used to refine the preliminary image description. As stated in \cite{gao2019deliberate}, the second-pass is based on global features captured by the visual attention and hidden layer in the first-pass. Hence, their proposed approach has the potential to generate a better image caption. In \cite{pan2020x}, the authors proposed a unified attention block that fully employs bilinear pooling to perform reasoning or selectively capitalize on visual information. Existing conventional medical image captioning methods, e.g., \cite{laserson2018textray,li2018hybrid,jing2018automatic}, are mainly based on traditional natural image captioning approaches. A limitation of traditional image captioning methods is that although these models work well on natural image captioning datasets, they do not generalize well to medical image captioning datasets. An image captioning model reinforced by context, e.g., keywords, is a more promising way to generate a better description for a medical image \cite{huang2021deepOpht}. In this work, a new context-driven model is introduced to improve the performance of the medical image captioning model.

\subsection{Multi-modal Medical Image Captioning}
Recently, a multi-modal task visual question answering (VQA) has been introduced \cite{malinowski2015ask,malinowski2014multi,malinowski2017ask,huang2019assessing,huang2018robustness,huang2017vqabq,huang2017robustness,huang2019novel}. The goal of VQA is to output a text-based answer for an input text-based question with a given image. Since a VQA model has textual and visual inputs, one modality is used to help the other \cite{agrawal2017vqa}. A similar idea of multiple input modalities can be also applied to build a multi-modal related \cite{jing2018automatic} or a multi-modal medical image captioning model \cite{huang2021deepOpht}. The authors of \cite{jing2018automatic} proposed a multi-modal related method to generate a medical description for a given lung X-ray image. Their proposed approach only has an image input modality and the image input is used to generate intermediate/side products, i.e., text-based tags, to reinforce the later generation of a medical description. However, the model-generated intermediate products could be wrong/bias information and it could confuse models during the training phase. Hence, in \cite{huang2021deepOpht}, a model with multi-modal inputs, i.e., a set of expert-defined keywords and a retina image, was introduced to generate a better quality of medical description. The expert-defined keywords help models learn better because the correctness/quality of the keywords is guaranteed by experts \cite{huang2021deepOpht}. Although considering expert-defined keywords as one of the multi-modal inputs could improve the model performance, effective textual input encoding will become another challenge. In this work, a new model with an effective textual input embedding method is introduced to tackle this challenge.

\subsection{Word Embeddings}
In this work, word embedding methods are categorized into three categories, i.e., the BoW, static word embeddings, and contextualized word representations.

\noindent\textbf{Bag of Words.} 
BoW model \cite{harris1954distributional} is a simplifying representation commonly used in information retrieval, NLP , and computer vision. In BoW, a sentence/document is represented as the bag of its words which keeps word multiplicity. The frequency of each word is used as a feature for training a model. According to \cite{scott1998text,soumya2014text,huang2021deepOpht}, since the BoW method likely cannot capture semantic meaning effectively, the method is not effective enough in multi-modal medical image captioning.

\noindent\textbf{Static Word Embeddings.} 
As stated in \cite{ethayarajh2019contextual}, Skip-gram with negative sampling (SGNS) \cite{mikolov2013distributed} and GloVe \cite{pennington2014glove} are among the best-known models for generating static word embeddings. In practice, these models learn word embeddings iteratively, but it has been proven that the models both implicitly factorize a word-context matrix containing a co-occurrence statistic \cite{levy2014neural,levy2014linguistic}. A notable issue with the static word embeddings is that all senses of a polysemous word must share a single vector because a single representation for each word is created.

\noindent\textbf{Contextualized Word Representations.} 
To tackle the above issue, the authors of \cite{peters2018deep,devlin2018bert,radford2019language} have proposed various deep neural language models to create context-sensitive word representations. The models are fine-tuned to create deep learning-based models for a wide range of downstream NLP tasks. In these models, the internal representations of words are considered as a function of the entire textual input. Hence, the representations are called contextualized word representations. In \cite{liu2019linguistic}, an approach was introduced suggesting that these representations capture task-agnostic and highly transferable properties of language. The authors of \cite{peters2018deep} introduced a method to generate contextualized representations of every token by concatenating the internal states of a 2-layer bi-LSTM. In \cite{radford2019language}, the proposed approach is an uni-directional transformer-based language model \cite{vaswani2017attention}. The method introduced in \cite{devlin2018bert} is a bi-directional transformer-based language model \cite{vaswani2017attention}. According to \cite{ethayarajh2019contextual}, the method of contextualized word representations is more effective than the static word embeddings. Hence, in this work, we propose to base on the contextualized word representations to develop a new model for multi-modal medical image captioning.

\subsection{Retinal Image Dataset}
Recently, various medical image datasets have been proposed for research, e.g., retinal image datasets \cite{hoover2003locating,staal2004ridge,kauppi2007diaretdb1,al2008reference,carmona2008identification,ortega2009retinal,niemeijer2011inspire,computing2012understanding,huang2021deep,fraz2012ensemble,huang2020query,vazquez2013improving,huang2021gpt2mvs,sivaswamy2014drishti,hu2019silco,yang2018novel,yang2018auto,liu2018synthesizing,decenciere2014feedback,adal2015accuracy,hernandez2017fire,porwal2018indian,huang2021deepOpht}. The above related existing retinal image datasets are reviewed in this subsection.

In \cite{hoover2003locating}, the proposed dataset STARE contains 397 images and it is used to develop an automatic system for diagnosing human eye diseases. The authors of \cite{staal2004ridge} proposed a dataset DRIVE with 40 retina images, half for the training set and the other half for the testing set. For the training, a single manual segmentation of the vasculature is available. For the testing, two manual segmentations are available. In \cite{kauppi2007diaretdb1}, the proposed dataset DIARETDB1 consists of 89 color fundus images. 84 images of them contain at least mild non-proliferative signs of Diabetic Retinopathy (DR). The remaining five images are considered normal without containing any signs of DR. In \cite{al2008reference}, the proposed dataset REVIEW with 14 images is designed for a segmentation task. The proposed dataset DRIONS-DB in \cite{carmona2008identification} contains 110 colorful retinal images. The proposed dataset contains several visual characteristics, e.g., cataract (severe or moderate), light artifacts, some of the rim blurred or missing, concentric peripapillary atrophy/artifacts, moderate peripapillary atrophy, and strong pallor distractor. The retinal image dataset VARIA \cite{ortega2009retinal} is used for authentication purposes. It contains 233 images from 139 different individuals. The proposed dataset INSPIRE-AVR in \cite{niemeijer2011inspire} consists of 40 colorful retinal images of the vessels and optic disc and an arterio-venous ratio reference standard. The authors of \cite{computing2012understanding} proposed a dataset ONHSD with 99 retinal images for a segmentation task. In \cite{fraz2012ensemble}, the introduced dataset CHASE-DB1 with 14 retinal images is used for retinal vessel analysis. The dataset VICAVR \cite{vazquez2013improving} for the computation of the ratio of Artery/Vein (A/V) consists of 58 retinal images. The proposed dataset Drishti-GS in \cite{sivaswamy2014drishti} has 101 images, 50 for training and 51 for testing. The MESSIDOR dataset \cite{decenciere2014feedback} has 1,200 colorful eye fundus images without manual annotation, e.g., lesions contours or position. The RODREP dataset \cite{adal2015accuracy} consists of repeated 1,120 4-field colorful fundus images of 70 patients in the DR screening program of the Rotterdam Eye Hospital. In \cite{hernandez2017fire}, the proposed dataset FIRE contains 129 retinal images, forming 134 image pairs. The image pairs are split into three different categories depending on their characteristics. The authors of \cite{porwal2018indian} proposed a dataset IDRiD containing 516 retinal fundus images. In the proposed dataset, the ground truths associated with the signs of DR, Diabetic Macular Edema (DME), and normal retinal structures are given as follows: (i) Pixel level labels of typical DR lesions and optic disc; (ii) Optic disc and fovea center coordinates; (iii) Image level disease severity grading of DR and DME. Although there are many existing retinal image datasets, not each of them is tailored for multi-modal retinal image captioning. Since the dataset introduced by \cite{huang2021deepOpht} is large-scale and specially designed for multi-modal deep learning research, in this work, the experiments are mainly based on \cite{huang2021deepOpht}'s DeepEyeNet (DEN) dataset.

\begin{figure*}
  \includegraphics[width=\textwidth]{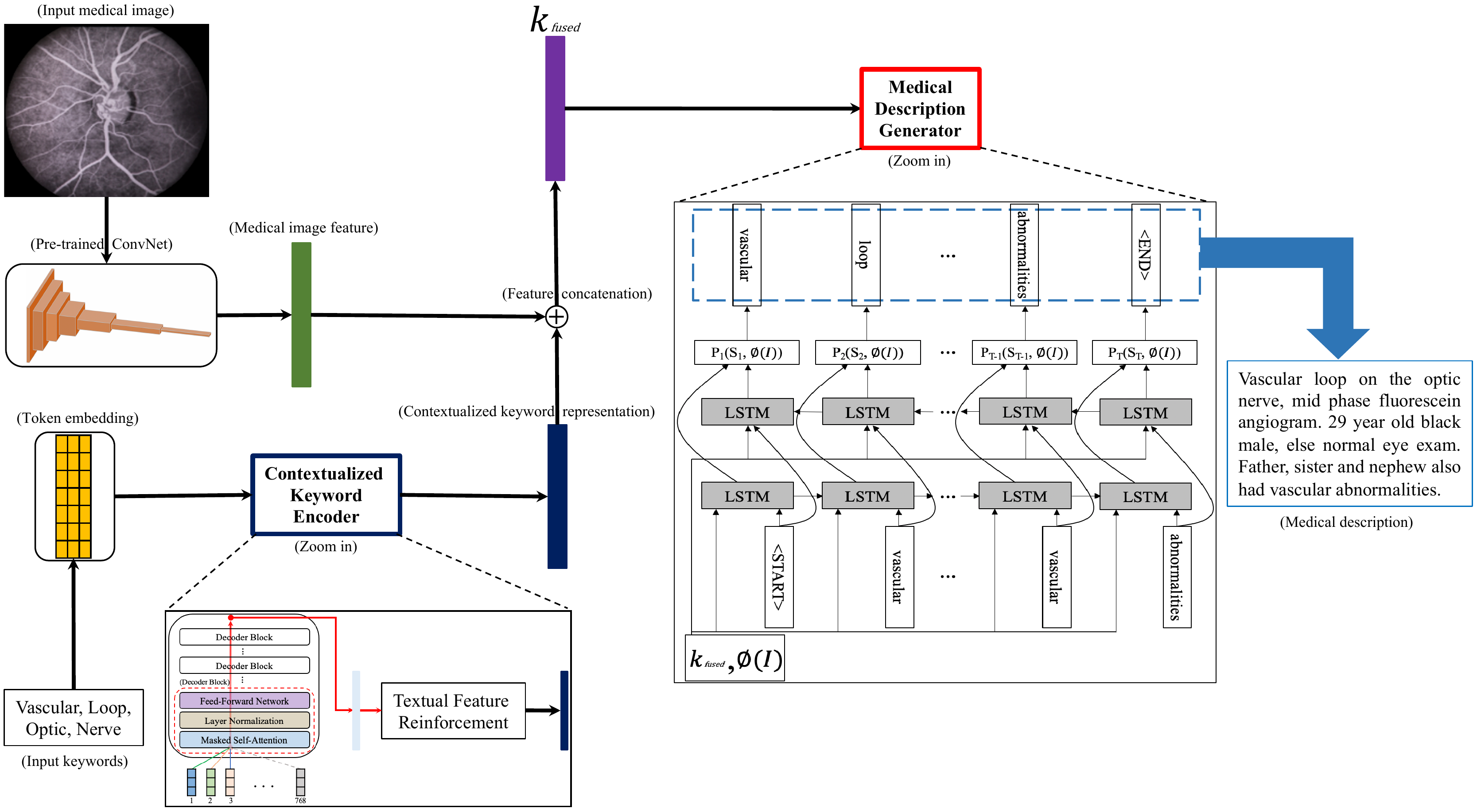}
\vspace{-0.8cm}
  \caption{
            Flowchart of proposed multi-modal medical image captioning model. A pre-trained CNN, e.g., VGG16 or VGG19 pre-trained on ImageNet, is used to extract features from the medical image input (dark green). From a set of input keywords, the ``Token embedding'' generates the input to the ``Contextualized Keyword Encoder'' which is composed of a stack of decoder blocks and ``Textual Feature Reinforcement''. Each decoder block consists of the masked self-attention, layer normalization, and feed-forward network (red dashed line box). ``Textual Feature Reinforcement'', i.e., a stack of fully-connected layers, generates the contextualized keyword representation (dark blue). Note that $768$ color-coded brick-stacked vectors are the input of the keyword encoder. $\oplus$ indicates the concatenation of the medical image feature and contextualized keyword representation. In ``Medical Description Generator'' which creates the medical description, $k_{fused}$ denotes a fused feature vector (purple), $\phi(I)$ denotes an image feature vector, and $P_{i}(S_{i}, \phi(I))$ is a probability distribution where $i=1,2,...,T$. See the \textit{Methodology} section for more details.
          }
  \label{fig:figure2}
\vspace{-0.1cm}
\end{figure*}


\section{Methodology}
In this section, the proposed multi-modal medical image captioning model is described in detail. The proposed method consists of a contextualized keyword encoder and a medical description generator. Flowchart of the proposed model is presented in Figure \ref{fig:figure2}.

\subsection{Contextualized Keyword Encoder}

Transformer consists of a transformer-encoder and a transformer-decoder \cite{vaswani2017attention}. Transformer architecture has been used with success in language modeling and machine translation. The transformer-encoder and transformer-decoder are the stacks of multiple basic transformer blocks. The proposed model is inspired by the GPT-2 structure, i.e., a transformer-decoder-like structure, in terms of its parallelization and masked self-attention. Its characteristics are deployed to develop the proposed contextualized keyword encoder for the embedding of keywords. Detail of the contextualized keyword encoder is presented as follows:
\begin{equation}
    x_n = W_e*k_n, n \in \{0,...,N-1\}, \\
	\label{eq:text_embed1}
\end{equation}
where $x_n$ is an input token/keyword embedding, $W_e \in \mathbb{R}^{E_s \times V_s}$ indicates the token embedding matrix, $E_s$ denotes the word embedding size, $V_s$ indicates the vocabulary size, $k_n$ denotes an input token/keyword, and $N$ denotes the number of input tokens/keywords.

\vspace{+0.1cm}\noindent\textbf{Masked Self-attention Mechanism.}
The mechanism of masked self-attention is described as follows:
\begin{equation}
    Q = W_q*x_n+b_q,
	\label{eq:text_embed2}
\end{equation}
where $Q$ denotes the representation of the current word \cite{vaswani2017attention}.  One linear layer, i.e., $W_q \in \mathbb{R}^{H_s \times E_s}$ with bias term $b_q$ and output size $H_s$, is used to generate $Q$.
\begin{equation}
    K = W_k*x_n+b_k,
	\label{eq:text_embed3}
\end{equation}
where $K$ is the key vector \cite{vaswani2017attention}. One linear layer, i.e., $W_k \in \mathbb{R}^{H_s \times E_s}$ with bias term $b_k$ and output size $H_s$, is used to generate $K$.
\begin{equation}
    V = W_v*x_n+b_v,
	\label{eq:text_embed4}
\end{equation}
where $V$ denotes the value vector \cite{vaswani2017attention}. One linear layer, i.e., $W_v \in \mathbb{R}^{H_s \times E_s}$ with bias term $b_v$ and output size $H_s$, is used to generate $V$.
\begin{equation}
    MaskAtten(Q,K,V) = softmax(m(\frac{QK^T}{\sqrt{d_k}}))V,
	\label{eq:attention}
\end{equation}
where $m(\cdot)$ is a masked self-attention function and $d_k$ denotes a scaling factor \cite{vaswani2017attention}.

\vspace{+0.1cm}\noindent\textbf{Layer Normalization.}
The layer normalization is calculated as Equation-(\ref{eq:layernorm}).
\begin{equation}
    Z_{Norm} = LayerNorm(MaskAtten(Q,K,V)),
	\label{eq:layernorm}
\end{equation}
where $LayerNorm(\cdot)$ is a function for layer normalization and $MaskAtten(Q,K,V)$ is the result from Equation-(\ref{eq:attention}).

\vspace{+0.1cm}\noindent\textbf{Contextualized Keyword Representation.}
Through the above, i.e., Equation-(\ref{eq:text_embed1}), Equation-(\ref{eq:text_embed2}), Equation-(\ref{eq:text_embed3}), Equation-(\ref{eq:text_embed4}), Equation-(\ref{eq:attention}), and Equation-(\ref{eq:layernorm}), the contextualized keyword representation $F$ is derived as:
\begin{equation}
    F = FFN(Z_{Norm}) = \sigma(W_1Z_{Norm}+b_1)W_2+b_2,
	\label{eq:ffn}
\end{equation}
where $FFN(\cdot)$ denotes a position-wise feed-forward network (FFN), $\sigma$ indicates an activation function. $W1$, $W2$, $b1$, and $b2$ are learnable parameters of the FFN. Note that in practice, a stack of fully-connected layers is used to reinforce $F$, referring to Figure \ref{fig:figure2}.

\subsection{Loss Function}

In this work, medical description generation is modeled as a classification problem. Hence, the categorical cross-entropy loss function is adopted to build the proposed method, referring to Equation-(\ref{eq:loss}).
\begin{equation}
    Loss = -\frac{1}{N}\sum_{i=0}^{N-1}\sum_{c=0}^{C-1}\mathbf{1}_{y_{i}\in C_{c}}\textup{log}(P_{model}\left [y_{i}\in C_{c} \right ]),
    \label{eq:loss}
\end{equation}
where $N$ denotes the number of observations, $C$ indicates the number of categories, $\mathbf{1}_{y_i \in C_c}$ denotes an indicator function of the $i$-th observation belonging to the $c$-th category, and $P_{model}[y_i \in C_c]$ is the probability predicted by the model for the $i$-th observation to belong to the $c$-th category. 

When there are more than two categories, the neural network model outputs a probability vector with $C$ dimensions. Each element in the vector gives the probability that the model input should be classified as belonging to the respective category. Note that when the number of categories is two, the categorical cross-entropy loss degenerates to the binary cross-entropy loss, i.e., a special case of the categorical cross-entropy loss. In this case, the neural network outputs a single probability $\hat{y}_i$, with the other one being $1-\hat{y}_i$. 
 
\subsection{Medical Description Generator}

For the medical description generator, the CNN medical image encoder $\phi$ used in \cite{huang2021deepOpht} is adopted to extract the image feature. The extracted feature is fed in each time step of a subsequent bidirectional LSTM model. $p(S_t|I, S_0,..., S_t-1)$ denotes all of the preceding words, and $S=(S_0,..., S_T)$ indicates a true sentence describing the input image $I$. 

The medical description generator is unrolled as follows:
\begin{equation}
    e_t = W_d\times \phi(I), t \in \{0,...,T\},
	\label{eq:4}
\end{equation}
where $W_d \in \mathbb{R}^{E \times F}$ denotes a fully-connected layer, $E$ represents the word embedding size, and $F$ is the image feature size.
\begin{equation}
    x_t = W_eS_t, t \in \{0,...,T\},
	\label{eq:5}
\end{equation}
where each word is represented as a bag-of-word id $S_t$, and the sentence $S$ and the image $I$ are mapped to the same high dimensional space.
\begin{equation}
    P_{t} = BiLSTM([e_t, k_{fused}, x_t]), t \in \{0,...,T\},
    \label{eq:6}
\end{equation}
where in Equation-(\ref{eq:6}), for each time step, image contents $e_t$, fused multi-modal feature $k_{fused}$, and ground truth word vector $x_t$ are fed to the network to strengthen its memory of images.

\begin{table*}
\caption{Comparison with the state-of-the-art ``DeepOpht with BoW'' \cite{huang2021deepOpht}, based on the metric of BLEU \cite{papineni2002bleu}, CIDEr \cite{vedantam2015cider}, ROUGE \cite{lin2004rouge}, and METEOR \cite{banerjee2005meteor}. Proposed method outperforms the model in \cite{huang2021deepOpht} by +$53.2$\% in BLEU-avg and +$18.6$\% in CIDEr. The results are based on VGG16 image feature extractor pre-trained on ImageNet and beam search algorithm in the testing phase. Note that BLEU-avg indicates the average score of BLEU-1, BLEU-2, BLEU-3, and BLEU-4. ''beam'' denotes number of beams used in the beam search. $*$ denotes not available. Similar notations also used in Table \ref{table:table2}. In general, the proposed method already beats baseline with beam=1, i.e., no beam search used, in BLEU and CIDEr and competitive in ROUGE-L.}
\vspace{-0.4cm}
\centering
\scalebox{1.0}{
\begin{tabular}{|c|c|c|c|c|c|c|c|c|c|}
\hline
Model                  & BLEU-1 & BLEU-2 & BLEU-3 & BLEU-4 & BLEU-avg & CIDEr & ROUGE-L & METEOR \\ \hline
DeepOpht with BoW (beam=3) \cite{huang2021deepOpht}  & 0.144  & 0.092  & 0.052  & 0.021  & 0.077  & 0.296 & \textbf{0.197} & $*$ \\ \cline{1-9}

Proposed Model with GloVe (beam=1)    & 0.173  & 0.111  & 0.072  & 0.048  & 0.101   & 0.243 & 0.164 & 0.153 \\ \cline{1-9} 
                                    
                                    
Proposed Model with GPT-2 (beam=1)   & \textbf{0.192}  & \textbf{0.130}  & \textbf{0.088}  & \textbf{0.060}  & \textbf{0.118}  & \textbf{0.351} & 0.188 & \textbf{0.167} \\ \cline{1-9} 
                                
                                 
\end{tabular}}
\label{table:table1}
\vspace{-0.1cm}
\end{table*}

\begin{table*}
\caption{Comparison with the state-of-the-art ``DeepOpht with BoW'' \cite{huang2021deepOpht}. The results are based on VGG19 image feature extractor pre-trained on ImageNet and beam search algorithm in the testing phase. Proposed method outperforms the model in \cite{huang2021deepOpht} by +$30$\% in BLEU-avg and +$8$\% in CIDEr.}
\vspace{-0.4cm}
\centering
\scalebox{1.0}{
\begin{tabular}{|c|c|c|c|c|c|c|c|c|c|}
\hline
Model                  & BLEU-1 & BLEU-2 & BLEU-3 & BLEU-4 & BLEU-avg & CIDEr & ROUGE-L & METEOR \\ \hline
DeepOpht with BoW (beam=3) \cite{huang2021deepOpht}     & 0.184  & 0.114  & 0.068  & 0.032  & 0.100  & 0.361 & \textbf{0.232} & $*$ \\ \cline{1-9}

Proposed Model with GloVe (beam=1)   & 0.192  & 0.132  & 0.093  & 0.067  & 0.121   & 0.356 & 0.186 & 0.169 \\ \cline{1-9} 
                                    
Proposed Model with GloVe (beam=3)   & 0.201  & 0.139  & 0.098  & 0.071  & 0.127  & 0.359 & 0.203 & 0.184 \\ \cline{1-9} 
                                    
Proposed Model with GPT-2 (beam=1)     & 0.203  & 0.137  & 0.093  & 0.065  & 0.125  & 0.356 & 0.197 & 0.174 \\ \cline{1-9} 
             
Proposed Model with GPT-2 (beam=3)     & \textbf{0.203}  & \textbf{0.142}  & \textbf{0.100}  & \textbf{0.073}  & \textbf{0.130}  & \textbf{0.389} & 0.211 & \textbf{0.188} \\ \cline{1-9}      
                                 
\end{tabular}}
\label{table:table2}
\end{table*}

\section{Experiments and Analysis}
In this section, the dataset and evaluation metrics used in the experiments are introduced and the experimental setup is described in detail. Then, the effectiveness of the proposed multi-modal medical image captioning model is analyzed. Finally, several randomly selected qualitative results are displayed.

\subsection{Dataset and Evaluation Metrics}

\noindent\textbf{Dataset.}
In \cite{huang2021deepOpht}, a state-of-the-art model and a large-scale retinal image dataset with unique expert-defined keyword annotations are introduced for multi-modal medical image captioning. The dataset is composed of $1,811$ grey-scale Fluorescein Angiography (FA) images and $13,898$ colorful Color Fundus Photography (CFP) images. Each image in the dataset has two corresponding labels, i.e., the clinical description and expert-defined keywords. In \cite{huang2021deepOpht}'s proposed dataset, the longest word length is more than $15$ words and $50$ words for keywords and clinical descriptions, respectively. The average word length of the keywords and clinical descriptions is between $5$ words and $10$ words. The dataset contains $265$ different retinal diseases/symptoms including the common and non-common. According to \cite{huang2021deepOpht}, the expert-defined keywords are collected from the ophthalmologists' or retinal specialists' retinal image analysis and diagnosis records with patients. Hence, the keywords contain information about potential retinal diseases, retinal symptoms, or patients' characteristics. In \cite{huang2021deepOpht}, the entire dataset is divided into $60\%/20\%/20\%$ for training/validation/testing, respectively.

\noindent\textbf{Evaluation Metrics.}
The same medical description evaluation metrics used in \cite{huang2021deepOpht} are adopted in this work to quantify the performance of the proposed model, i.e., BLEU \cite{papineni2002bleu}, CIDEr \cite{vedantam2015cider}, and ROUGE \cite{lin2004rouge}. Another commonly used text evaluation metric METEOR \cite{banerjee2005meteor} is also used to evaluate the proposed method. The aforementioned evaluation metrics are defined as follows: 
\begin{equation}
    \textup{BP}= \left\{\begin{matrix}
                        1 & \textup{if} & c>r \\ 
                        \textup{exp}(1-\frac{r}{c}) & \textup{if} & c\leq r
                        \end{matrix}\right.;
    \textup{BLEU}= \textup{BP}\cdot \textup{exp}\left ( \sum_{n=1}^{N} w_{n}\textup{log}p_{n} \right ),
    \label{eq:bleu}
\end{equation}
where $r$ denotes the effective ground truth text length, $c$ indicates the length of the prediction text, and $\textup{BP}$ denotes brevity penalty. The geometric average of the modified $n$-gram precisions $p_{n}$ is computed by using $n$-grams up to length $N$ and positive weights $w_{n}$ summing to $1$.

\begin{equation}
    \begin{split}
    \textup{CIDEr}_{n}(c_{i},S_{i})= \frac{1}{m}\sum_{j}\frac{\boldsymbol{g}^{\boldsymbol{n}}(c_{i})\cdot               
        \boldsymbol{g}^{\boldsymbol{n}}(s_{ij})}{\left \| \boldsymbol{g}^{\boldsymbol{n}}(c_{i}) \right \|\left \| \boldsymbol{g}^{\boldsymbol{n}}(s_{ij}) \right \|};\\
    \textup{CIDEr}(c_{i},S_{i})= \sum_{n=1}^{N}w_{n}\textup{CIDEr}_{n}(c_{i},S_{i}),
    \end{split}
    \label{eq:f1-cider}
\end{equation}
where $c_{i}$ denotes prediction text, $S_{i}$ = \{$s_{i1}$,..., $s_{im}$\} denotes a set of ground truth descriptions. $\textup{CIDEr}_{n}(c_{i},S_{i})$ score for $n$-grams of length $n$ is computed by using the average cosine similarity between $c_{i}$ and $S_{i}$, which accounts for both precision and recall. $\boldsymbol{g}^{\boldsymbol{n}}(c_{i})$ is a vector formed by $g_{k}(c_{i})$ corresponding to all $n$-grams of length $n$, and $\left \| \boldsymbol{g}^{\boldsymbol{n}}(c_{i}) \right \|$ denotes the magnitude of $\boldsymbol{g}^{\boldsymbol{n}}(c_{i})$. Similarly for $\boldsymbol{g}^{\boldsymbol{n}}(s_{ij})$. The higher order (longer) $n$-grams is used to capture grammatical properties as well as richer semantics. $\textup{CIDEr}(c_{i},S_{i})$ indicated the combined score based on $n$-grams of varying lengths.

\begin{equation}
    \textup{R}_{lcs}=\frac{\textup{LCS}(X,Y)}{m};
    \textup{P}_{lcs}=\frac{\textup{LCS}(X,Y)}{n};
    \textup{F}_{lcs}=\frac{(1+\beta^{2})\textup{R}_{lcs}\textup{P}_{lcs}}{\textup{R}_{lcs}+\beta^{2}\textup{P}_{lcs}},
    \label{eq:f1-rouge}
\end{equation}
where the longest common subsequence (LCS) based $\textup{F}$-measure $\textup{F}_{lcs}$/ROUGE-L is used to estimate the similarity between ground truth text $X$ = \{$x_{1}$,..., $x_{m}$\} with length $m$ and prediction text $Y$ = \{$y_{1}$,..., $y_{n}$\} with length $n$. $\beta$ is to balance the relative importance between $\textup{P}_{lcs}$ and $\textup{R}_{lcs}$.

\begin{equation}
    \textup{METEOR}_{score}=\frac{10PR}{R+9P}\left ( 1-0.5(\frac{\#chunks}{\#unigrams\_matched})^{3} \right ),
    \label{eq:f1-meteor}
\end{equation}
where $P$ denotes unigram precision, $R$ denotes unigram recall, $\#chunks$ indicates number of chunks, and $\#unigrams\_matched$ denotes number of matched unigrams, referring to \cite{banerjee2005meteor} for detail.

\begin{figure}[ht]
\includegraphics[width=1.0\linewidth]{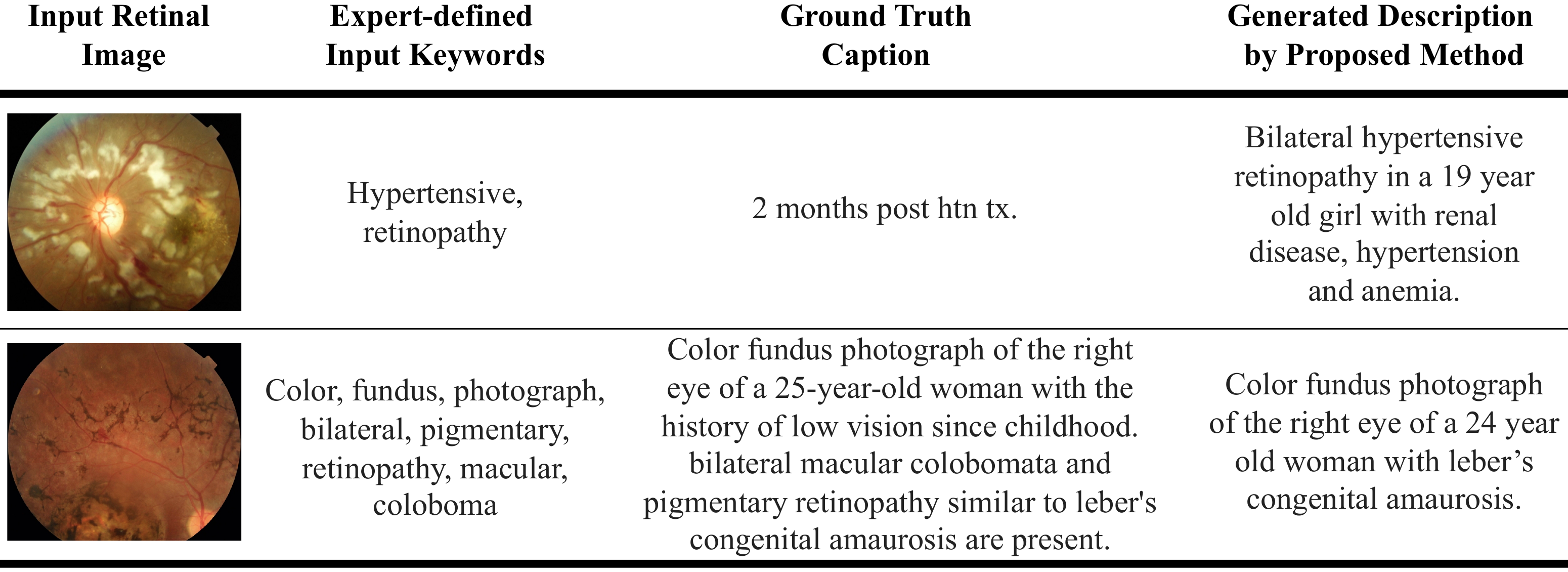}
\vspace{-0.8cm}
\caption{
Randomly selected qualitative results of the proposed multi-modal medical image captioning. The result shows that the proposed model exploits the effective contextualized keyword representations as guidance to generate meaningful medical descriptions. Note that in practice, ``date'', ``skin color'', ``gender'', and ``age'' would be part of the dataset and that a system should make it part of the medical description by slot filling or post-processing \cite{huang2021deepOpht}.
}
\label{fig:figure3}
\vspace{-0.3cm}
\end{figure}

\subsection{Experimental Setup}
Similar to \cite{huang2021deepOpht}, in this work, VGG16 and VGG19 pre-trained on ImageNet are adopted to extract image features. To process keywords and descriptions, non-alphabet characters are removed, all remaining characters are converted to lower-case, and all the words appearing only once are replaced by a special token $<UNK>$. When keywords are excluded, the vocabulary size is $4,007$. When keywords are included, the vocabulary size is $4,292$. All sentences are truncated or padded with a maximum length of $50$. A token/word embedding size $300$ is used to encode every word. Since the proposed contextualized keyword encoder is based on the GPT-2 architecture \cite{radford2019language}, using the pre-trained weight of GPT-2 for initialization is helpful in the experiments. As stated in \cite{radford2019language}, GPT-2 has been pre-trained on a large corpus with vocabulary size $50,257$. For the proposed medical description generator, a hidden layer size $256$ is used for the LSTM unit. The default setup of \cite{huang2021deepOpht}'s dataset is used for experiments, i.e., $60\%/20\%/20\%$ for training/validation/testing, respectively. In this work, Keras is used for implementation, and models are trained with $2$ epochs, $64$ batch size, $1e-3$ learning rate, and Adam optimizer \cite{kingma2014adam}. For the hyperparameters of Adam optimizer, coefficients used for computing moving averages of gradient and its square are $\beta_{1}=0.9$ and $\beta_{2}=0.999$, respectively. $\epsilon=1e-8$ is added to the denominator to improve numerical stability.

\subsection{Effectiveness Analysis}
According to Table \ref{table:table1} and Table \ref{table:table2}, the results show that the proposed model with contextualized keyword representations and static word embeddings beat the baseline model with BoW \cite{huang2021deepOpht}. Moreover, the model with contextualized keyword representations outperforms the model with static word embeddings \cite{pennington2014glove}. Since contextualized keyword representations effectively capture the keywords information, the performance of the multi-modal medical image captioning model is improved, +$53.2$\% in BLEU-avg and +$18.6$\% in CIDEr, and a better quality of medical description is generated. Note that although the beam search algorithm with three beams boosts model performance in the testing phase, the computational cost is around $12$ times of with one beam ($14851/1265$ in seconds). Qualitative results are demonstrated in Figure \ref{fig:figure3}.

\section{Conclusion}
To sum up, in this paper a new end-to-end deep model is introduced for multi-modal medical image captioning. The contextualized keyword representation, textual feature reinforcement module, and masked self-attention are used to develop the proposed method. The effectiveness of the proposed model is thoroughly evaluated through experiments on the existing multi-modal retinal image captioning dataset proposed by \cite{huang2021deepOpht}. The experimental results show that the proposed method outperforms the baseline model. The model performance is increased in terms of BLEU and CIDEr.

\begin{acks}
This work is supported by competitive research funding from the University of Amsterdam.
\end{acks}

%
\bibliographystyle{ACM-Reference-Format}
\bibliography{sample-base}


\begin{thebibliography}{62}


\ifx \showCODEN    \undefined \def \showCODEN     #1{\unskip}     \fi
\ifx \showDOI      \undefined \def \showDOI       #1{#1}\fi
\ifx \showISBNx    \undefined \def \showISBNx     #1{\unskip}     \fi
\ifx \showISBNxiii \undefined \def \showISBNxiii  #1{\unskip}     \fi
\ifx \showISSN     \undefined \def \showISSN      #1{\unskip}     \fi
\ifx \showLCCN     \undefined \def \showLCCN      #1{\unskip}     \fi
\ifx \shownote     \undefined \def \shownote      #1{#1}          \fi
\ifx \showarticletitle \undefined \def \showarticletitle #1{#1}   \fi
\ifx \showURL      \undefined \def \showURL       {\relax}        \fi
\providecommand\bibfield[2]{#2}
\providecommand\bibinfo[2]{#2}
\providecommand\natexlab[1]{#1}
\providecommand\showeprint[2][]{arXiv:#2}

\bibitem[\protect\citeauthoryear{Adal, van Etten, Martinez, van Vliet, and
  Vermeer}{Adal et~al\mbox{.}}{2015}]%
        {adal2015accuracy}
\bibfield{author}{\bibinfo{person}{Kedir~M Adal}, \bibinfo{person}{Peter~G van
  Etten}, \bibinfo{person}{Jose~P Martinez}, \bibinfo{person}{Lucas~J van
  Vliet}, {and} \bibinfo{person}{Koenraad~A Vermeer}.}
  \bibinfo{year}{2015}\natexlab{}.
\newblock \showarticletitle{Accuracy assessment of intra-and intervisit fundus
  image registration for diabetic retinopathy screening}.
\newblock \bibinfo{journal}{\emph{Investigative ophthalmology \& visual
  science}} \bibinfo{volume}{56}, \bibinfo{number}{3} (\bibinfo{year}{2015}),
  \bibinfo{pages}{1805--1812}.
\newblock


\bibitem[\protect\citeauthoryear{Agrawal, Lu, Antol, Mitchell, Zitnick, Parikh,
  and Batra}{Agrawal et~al\mbox{.}}{2017}]%
        {agrawal2017vqa}
\bibfield{author}{\bibinfo{person}{Aishwarya Agrawal}, \bibinfo{person}{Jiasen
  Lu}, \bibinfo{person}{Stanislaw Antol}, \bibinfo{person}{Margaret Mitchell},
  \bibinfo{person}{C~Lawrence Zitnick}, \bibinfo{person}{Devi Parikh}, {and}
  \bibinfo{person}{Dhruv Batra}.} \bibinfo{year}{2017}\natexlab{}.
\newblock \showarticletitle{Vqa: Visual question answering}.
\newblock \bibinfo{journal}{\emph{International Journal of Computer Vision}}
  \bibinfo{volume}{123}, \bibinfo{number}{1} (\bibinfo{year}{2017}),
  \bibinfo{pages}{4--31}.
\newblock


\bibitem[\protect\citeauthoryear{Al-Diri, Hunter, Steel, Habib, Hudaib, and
  Berry}{Al-Diri et~al\mbox{.}}{2008}]%
        {al2008reference}
\bibfield{author}{\bibinfo{person}{Bashir Al-Diri}, \bibinfo{person}{Andrew
  Hunter}, \bibinfo{person}{David Steel}, \bibinfo{person}{Maged Habib},
  \bibinfo{person}{Taghread Hudaib}, {and} \bibinfo{person}{Simon Berry}.}
  \bibinfo{year}{2008}\natexlab{}.
\newblock \showarticletitle{A reference data set for retinal vessel profiles}.
  In \bibinfo{booktitle}{\emph{2008 30th Annual International Conference of the
  IEEE Engineering in Medicine and Biology Society}}. IEEE,
  \bibinfo{pages}{2262--2265}.
\newblock


\bibitem[\protect\citeauthoryear{Banerjee and Lavie}{Banerjee and
  Lavie}{2005}]%
        {banerjee2005meteor}
\bibfield{author}{\bibinfo{person}{Satanjeev Banerjee} {and}
  \bibinfo{person}{Alon Lavie}.} \bibinfo{year}{2005}\natexlab{}.
\newblock \showarticletitle{METEOR: An automatic metric for MT evaluation with
  improved correlation with human judgments}. In
  \bibinfo{booktitle}{\emph{Proceedings of the acl workshop on intrinsic and
  extrinsic evaluation measures for machine translation and/or summarization}}.
  \bibinfo{pages}{65--72}.
\newblock


\bibitem[\protect\citeauthoryear{Carmona, Rinc{\'o}n, Garc{\'\i}a-Feijo{\'o},
  and Mart{\'\i}nez-de-la Casa}{Carmona et~al\mbox{.}}{2008}]%
        {carmona2008identification}
\bibfield{author}{\bibinfo{person}{Enrique~J Carmona}, \bibinfo{person}{Mariano
  Rinc{\'o}n}, \bibinfo{person}{Juli{\'a}n Garc{\'\i}a-Feijo{\'o}}, {and}
  \bibinfo{person}{Jos{\'e}~M Mart{\'\i}nez-de-la Casa}.}
  \bibinfo{year}{2008}\natexlab{}.
\newblock \showarticletitle{Identification of the optic nerve head with genetic
  algorithms}.
\newblock \bibinfo{journal}{\emph{Artificial Intelligence in Medicine}}
  \bibinfo{volume}{43}, \bibinfo{number}{3} (\bibinfo{year}{2008}),
  \bibinfo{pages}{243--259}.
\newblock


\bibitem[\protect\citeauthoryear{Computing}{Computing}{2012}]%
        {computing2012understanding}
\bibfield{author}{\bibinfo{person}{Retinal~Image Computing}.}
  \bibinfo{year}{2012}\natexlab{}.
\newblock \bibinfo{title}{Understanding,“ONHSD-Optic Nerve Head Segmentation
  Dataset,” University of Lincoln, United Kingdom, 2004}.
\newblock
\newblock


\bibitem[\protect\citeauthoryear{Decenci{\`e}re, Zhang, Cazuguel, Lay,
  Cochener, Trone, Gain, Ordonez, Massin, Erginay,
  et~al\mbox{.}}{Decenci{\`e}re et~al\mbox{.}}{2014}]%
        {decenciere2014feedback}
\bibfield{author}{\bibinfo{person}{Etienne Decenci{\`e}re},
  \bibinfo{person}{Xiwei Zhang}, \bibinfo{person}{Guy Cazuguel},
  \bibinfo{person}{Bruno Lay}, \bibinfo{person}{B{\'e}atrice Cochener},
  \bibinfo{person}{Caroline Trone}, \bibinfo{person}{Philippe Gain},
  \bibinfo{person}{Richard Ordonez}, \bibinfo{person}{Pascale Massin},
  \bibinfo{person}{Ali Erginay}, {et~al\mbox{.}}}
  \bibinfo{year}{2014}\natexlab{}.
\newblock \showarticletitle{Feedback on a publicly distributed image database:
  the Messidor database}.
\newblock \bibinfo{journal}{\emph{Image Analysis \& Stereology}}
  \bibinfo{volume}{33}, \bibinfo{number}{3} (\bibinfo{year}{2014}),
  \bibinfo{pages}{231--234}.
\newblock


\bibitem[\protect\citeauthoryear{Devlin, Chang, Lee, and Toutanova}{Devlin
  et~al\mbox{.}}{2018}]%
        {devlin2018bert}
\bibfield{author}{\bibinfo{person}{Jacob Devlin}, \bibinfo{person}{Ming-Wei
  Chang}, \bibinfo{person}{Kenton Lee}, {and} \bibinfo{person}{Kristina
  Toutanova}.} \bibinfo{year}{2018}\natexlab{}.
\newblock \showarticletitle{Bert: Pre-training of deep bidirectional
  transformers for language understanding}.
\newblock \bibinfo{journal}{\emph{arXiv preprint arXiv:1810.04805}}
  (\bibinfo{year}{2018}).
\newblock


\bibitem[\protect\citeauthoryear{Ethayarajh}{Ethayarajh}{2019}]%
        {ethayarajh2019contextual}
\bibfield{author}{\bibinfo{person}{Kawin Ethayarajh}.}
  \bibinfo{year}{2019}\natexlab{}.
\newblock \showarticletitle{How contextual are contextualized word
  representations? comparing the geometry of BERT, ELMo, and GPT-2 embeddings}.
\newblock \bibinfo{journal}{\emph{arXiv preprint arXiv:1909.00512}}
  (\bibinfo{year}{2019}).
\newblock


\bibitem[\protect\citeauthoryear{Fang, Gupta, Iandola, Srivastava, Deng,
  Doll{\'a}r, Gao, He, Mitchell, Platt, et~al\mbox{.}}{Fang
  et~al\mbox{.}}{2015}]%
        {fang2015captions}
\bibfield{author}{\bibinfo{person}{Hao Fang}, \bibinfo{person}{Saurabh Gupta},
  \bibinfo{person}{Forrest Iandola}, \bibinfo{person}{Rupesh~K Srivastava},
  \bibinfo{person}{Li Deng}, \bibinfo{person}{Piotr Doll{\'a}r},
  \bibinfo{person}{Jianfeng Gao}, \bibinfo{person}{Xiaodong He},
  \bibinfo{person}{Margaret Mitchell}, \bibinfo{person}{John~C Platt},
  {et~al\mbox{.}}} \bibinfo{year}{2015}\natexlab{}.
\newblock \showarticletitle{From captions to visual concepts and back}. In
  \bibinfo{booktitle}{\emph{Proceedings of the IEEE conference on computer
  vision and pattern recognition}}. \bibinfo{pages}{1473--1482}.
\newblock


\bibitem[\protect\citeauthoryear{Fraz, Remagnino, Hoppe, Uyyanonvara, Rudnicka,
  Owen, and Barman}{Fraz et~al\mbox{.}}{2012}]%
        {fraz2012ensemble}
\bibfield{author}{\bibinfo{person}{Muhammad~Moazam Fraz},
  \bibinfo{person}{Paolo Remagnino}, \bibinfo{person}{Andreas Hoppe},
  \bibinfo{person}{Bunyarit Uyyanonvara}, \bibinfo{person}{Alicja~R Rudnicka},
  \bibinfo{person}{Christopher~G Owen}, {and} \bibinfo{person}{Sarah~A
  Barman}.} \bibinfo{year}{2012}\natexlab{}.
\newblock \showarticletitle{An ensemble classification-based approach applied
  to retinal blood vessel segmentation}.
\newblock \bibinfo{journal}{\emph{IEEE Transactions on Biomedical Engineering}}
  \bibinfo{volume}{59}, \bibinfo{number}{9} (\bibinfo{year}{2012}),
  \bibinfo{pages}{2538--2548}.
\newblock


\bibitem[\protect\citeauthoryear{Gao, Fan, Song, Liu, Xu, and Shen}{Gao
  et~al\mbox{.}}{2019}]%
        {gao2019deliberate}
\bibfield{author}{\bibinfo{person}{Lianli Gao}, \bibinfo{person}{Kaixuan Fan},
  \bibinfo{person}{Jingkuan Song}, \bibinfo{person}{Xianglong Liu},
  \bibinfo{person}{Xing Xu}, {and} \bibinfo{person}{Heng~Tao Shen}.}
  \bibinfo{year}{2019}\natexlab{}.
\newblock \showarticletitle{Deliberate Attention Networks for Image
  Captioning}.
\newblock \bibinfo{journal}{\emph{AAAI}} (\bibinfo{year}{2019}).
\newblock


\bibitem[\protect\citeauthoryear{Harris}{Harris}{1954}]%
        {harris1954distributional}
\bibfield{author}{\bibinfo{person}{Zellig~S Harris}.}
  \bibinfo{year}{1954}\natexlab{}.
\newblock \showarticletitle{Distributional structure}.
\newblock \bibinfo{journal}{\emph{Word}} \bibinfo{volume}{10},
  \bibinfo{number}{2-3} (\bibinfo{year}{1954}), \bibinfo{pages}{146--162}.
\newblock


\bibitem[\protect\citeauthoryear{Hendricks, Akata, Rohrbach, Donahue, Schiele,
  and Darrell}{Hendricks et~al\mbox{.}}{2016}]%
        {hendricks2016generating}
\bibfield{author}{\bibinfo{person}{Lisa~Anne Hendricks},
  \bibinfo{person}{Zeynep Akata}, \bibinfo{person}{Marcus Rohrbach},
  \bibinfo{person}{Jeff Donahue}, \bibinfo{person}{Bernt Schiele}, {and}
  \bibinfo{person}{Trevor Darrell}.} \bibinfo{year}{2016}\natexlab{}.
\newblock \showarticletitle{Generating visual explanations}. In
  \bibinfo{booktitle}{\emph{European Conference on Computer Vision}}. Springer,
  \bibinfo{pages}{3--19}.
\newblock


\bibitem[\protect\citeauthoryear{Hernandez-Matas, Zabulis, Triantafyllou,
  Anyfanti, Douma, and Argyros}{Hernandez-Matas et~al\mbox{.}}{2017}]%
        {hernandez2017fire}
\bibfield{author}{\bibinfo{person}{Carlos Hernandez-Matas},
  \bibinfo{person}{Xenophon Zabulis}, \bibinfo{person}{Areti Triantafyllou},
  \bibinfo{person}{Panagiota Anyfanti}, \bibinfo{person}{Stella Douma}, {and}
  \bibinfo{person}{Antonis~A Argyros}.} \bibinfo{year}{2017}\natexlab{}.
\newblock \showarticletitle{FIRE: fundus image registration dataset}.
\newblock \bibinfo{journal}{\emph{Journal for Modeling in Ophthalmology}}
  \bibinfo{volume}{1}, \bibinfo{number}{4} (\bibinfo{year}{2017}),
  \bibinfo{pages}{16--28}.
\newblock


\bibitem[\protect\citeauthoryear{Hoover and Goldbaum}{Hoover and
  Goldbaum}{2003}]%
        {hoover2003locating}
\bibfield{author}{\bibinfo{person}{Adam Hoover} {and} \bibinfo{person}{Michael
  Goldbaum}.} \bibinfo{year}{2003}\natexlab{}.
\newblock \showarticletitle{Locating the optic nerve in a retinal image using
  the fuzzy convergence of the blood vessels}.
\newblock \bibinfo{journal}{\emph{IEEE transactions on medical imaging}}
  \bibinfo{volume}{22}, \bibinfo{number}{8} (\bibinfo{year}{2003}),
  \bibinfo{pages}{951--958}.
\newblock


\bibitem[\protect\citeauthoryear{Hu, Mettes, Huang, and Snoek}{Hu
  et~al\mbox{.}}{2019}]%
        {hu2019silco}
\bibfield{author}{\bibinfo{person}{Tao Hu}, \bibinfo{person}{Pascal Mettes},
  \bibinfo{person}{Jia-Hong Huang}, {and} \bibinfo{person}{Cees~GM Snoek}.}
  \bibinfo{year}{2019}\natexlab{}.
\newblock \showarticletitle{Silco: Show a few images, localize the common
  object}. In \bibinfo{booktitle}{\emph{Proceedings of the IEEE/CVF
  International Conference on Computer Vision}}. \bibinfo{pages}{5067--5076}.
\newblock


\bibitem[\protect\citeauthoryear{Huang}{Huang}{2017}]%
        {huang2017robustness}
\bibfield{author}{\bibinfo{person}{Jia-Hong Huang}.}
  \bibinfo{year}{2017}\natexlab{}.
\newblock \showarticletitle{Robustness Analysis of Visual Question Answering
  Models by Basic Questions}.
\newblock \bibinfo{journal}{\emph{King Abdullah University of Science and
  Technology MS thesis}} (\bibinfo{year}{2017}).
\newblock


\bibitem[\protect\citeauthoryear{Huang, Alfadly, and Ghanem}{Huang
  et~al\mbox{.}}{2017}]%
        {huang2017vqabq}
\bibfield{author}{\bibinfo{person}{Jia-Hong Huang}, \bibinfo{person}{Modar
  Alfadly}, {and} \bibinfo{person}{Bernard Ghanem}.}
  \bibinfo{year}{2017}\natexlab{}.
\newblock \showarticletitle{Vqabq: Visual question answering by basic
  questions}.
\newblock \bibinfo{journal}{\emph{CVPR VQA Challenge Workshop}}
  (\bibinfo{year}{2017}).
\newblock


\bibitem[\protect\citeauthoryear{Huang, Alfadly, Ghanem, and Worring}{Huang
  et~al\mbox{.}}{2019a}]%
        {huang2019assessing}
\bibfield{author}{\bibinfo{person}{Jia-Hong Huang}, \bibinfo{person}{Modar
  Alfadly}, \bibinfo{person}{Bernard Ghanem}, {and} \bibinfo{person}{Marcel
  Worring}.} \bibinfo{year}{2019}\natexlab{a}.
\newblock \showarticletitle{Assessing the robustness of visual question
  answering}.
\newblock \bibinfo{journal}{\emph{arXiv preprint arXiv:1912.01452}}
  (\bibinfo{year}{2019}).
\newblock


\bibitem[\protect\citeauthoryear{Huang, Dao, Alfadly, and Ghanem}{Huang
  et~al\mbox{.}}{2019b}]%
        {huang2019novel}
\bibfield{author}{\bibinfo{person}{Jia-Hong Huang}, \bibinfo{person}{Cuong~Duc
  Dao}, \bibinfo{person}{Modar Alfadly}, {and} \bibinfo{person}{Bernard
  Ghanem}.} \bibinfo{year}{2019}\natexlab{b}.
\newblock \showarticletitle{A novel framework for robustness analysis of visual
  qa models}. In \bibinfo{booktitle}{\emph{Proceedings of the AAAI Conference
  on Artificial Intelligence}}, Vol.~\bibinfo{volume}{33}.
  \bibinfo{pages}{8449--8456}.
\newblock


\bibitem[\protect\citeauthoryear{Huang, Dao, Alfadly, Yang, and Ghanem}{Huang
  et~al\mbox{.}}{2018}]%
        {huang2018robustness}
\bibfield{author}{\bibinfo{person}{Jia-Hong Huang}, \bibinfo{person}{Cuong~Duc
  Dao}, \bibinfo{person}{Modar Alfadly}, \bibinfo{person}{C~Huck Yang}, {and}
  \bibinfo{person}{Bernard Ghanem}.} \bibinfo{year}{2018}\natexlab{}.
\newblock \showarticletitle{Robustness analysis of visual qa models by basic
  questions}.
\newblock \bibinfo{journal}{\emph{CVPR VQA Challenge and Visual Dialog
  Workshop}} (\bibinfo{year}{2018}).
\newblock


\bibitem[\protect\citeauthoryear{Huang, Murn, Mrak, and Worring}{Huang
  et~al\mbox{.}}{2021a}]%
        {huang2021gpt2mvs}
\bibfield{author}{\bibinfo{person}{Jia-Hong Huang}, \bibinfo{person}{Luka
  Murn}, \bibinfo{person}{Marta Mrak}, {and} \bibinfo{person}{Marcel Worring}.}
  \bibinfo{year}{2021}\natexlab{a}.
\newblock \showarticletitle{GPT2MVS: Generative Pre-trained Transformer-2
  forMulti-modal Video Summarization}. In \bibinfo{booktitle}{\emph{Proceedings
  of the 2021 International Conference on Multimedia Retrieval}}.
  \bibinfo{pages}{242--250}.
\newblock


\bibitem[\protect\citeauthoryear{Huang and Worring}{Huang and Worring}{2020}]%
        {huang2020query}
\bibfield{author}{\bibinfo{person}{Jia-Hong Huang} {and}
  \bibinfo{person}{Marcel Worring}.} \bibinfo{year}{2020}\natexlab{}.
\newblock \showarticletitle{Query-controllable video summarization}. In
  \bibinfo{booktitle}{\emph{Proceedings of the 2020 International Conference on
  Multimedia Retrieval}}. \bibinfo{pages}{242--250}.
\newblock


\bibitem[\protect\citeauthoryear{Huang, Wu, Yang, and Worring}{Huang
  et~al\mbox{.}}{2021b}]%
        {huang2021deep}
\bibfield{author}{\bibinfo{person}{Jia-Hong Huang}, \bibinfo{person}{Ting-Wei
  Wu}, \bibinfo{person}{Chao-Han~Huck Yang}, {and} \bibinfo{person}{Marcel
  Worring}.} \bibinfo{year}{2021}\natexlab{b}.
\newblock \showarticletitle{Deep Context-Encoding Network for Retinal Image
  Captioning}.
\newblock \bibinfo{journal}{\emph{arXiv preprint arXiv:}}.
\newblock


\bibitem[\protect\citeauthoryear{Huang, Yang, Liu, Tian, Liu, Wu, Lin, Wang,
  Morikawa, Chang, et~al\mbox{.}}{Huang et~al\mbox{.}}{2021c}]%
        {huang2021deepOpht}
\bibfield{author}{\bibinfo{person}{Jia-Hong Huang}, \bibinfo{person}{C-H~Huck
  Yang}, \bibinfo{person}{Fangyu Liu}, \bibinfo{person}{Meng Tian},
  \bibinfo{person}{Yi-Chieh Liu}, \bibinfo{person}{Ting-Wei Wu},
  \bibinfo{person}{I Lin}, \bibinfo{person}{Kang Wang},
  \bibinfo{person}{Hiromasa Morikawa}, \bibinfo{person}{Hernghua Chang},
  {et~al\mbox{.}}} \bibinfo{year}{2021}\natexlab{c}.
\newblock \showarticletitle{DeepOpht: medical report generation for retinal
  images via deep models and visual explanation}. In
  \bibinfo{booktitle}{\emph{Proceedings of the IEEE/CVF winter conference on
  applications of computer vision}}. \bibinfo{pages}{2442--2452}.
\newblock


\bibitem[\protect\citeauthoryear{Jing, Xie, Xing, Jing, Xie, and Xing}{Jing
  et~al\mbox{.}}{2018}]%
        {jing2018automatic}
\bibfield{author}{\bibinfo{person}{Baoyu Jing}, \bibinfo{person}{Pengtao Xie},
  \bibinfo{person}{Eric Xing}, \bibinfo{person}{Baoyu Jing},
  \bibinfo{person}{Pengtao Xie}, {and} \bibinfo{person}{Eric Xing}.}
  \bibinfo{year}{2018}\natexlab{}.
\newblock \showarticletitle{On the automatic generation of medical imaging
  reports}.
\newblock \bibinfo{journal}{\emph{ACL}} (\bibinfo{year}{2018}).
\newblock


\bibitem[\protect\citeauthoryear{Karpathy and Fei-Fei}{Karpathy and
  Fei-Fei}{2015}]%
        {karpathy2015deep}
\bibfield{author}{\bibinfo{person}{Andrej Karpathy} {and} \bibinfo{person}{Li
  Fei-Fei}.} \bibinfo{year}{2015}\natexlab{}.
\newblock \showarticletitle{Deep visual-semantic alignments for generating
  image descriptions}. In \bibinfo{booktitle}{\emph{CVPR}}.
  \bibinfo{pages}{3128--3137}.
\newblock


\bibitem[\protect\citeauthoryear{Kauppi, Kalesnykiene, Kamarainen, Lensu,
  Sorri, Raninen, Voutilainen, Pietil{\"a}, K{\"a}lvi{\"a}inen, and
  Uusitalo}{Kauppi et~al\mbox{.}}{2007}]%
        {kauppi2007diaretdb1}
\bibfield{author}{\bibinfo{person}{Tomi Kauppi}, \bibinfo{person}{Valentina
  Kalesnykiene}, \bibinfo{person}{Joni-Kristian Kamarainen},
  \bibinfo{person}{Lasse Lensu}, \bibinfo{person}{Iiris Sorri},
  \bibinfo{person}{A Raninen}, \bibinfo{person}{R Voutilainen},
  \bibinfo{person}{J Pietil{\"a}}, \bibinfo{person}{H K{\"a}lvi{\"a}inen},
  {and} \bibinfo{person}{H Uusitalo}.} \bibinfo{year}{2007}\natexlab{}.
\newblock \bibinfo{title}{DIARETDB1—Standard Diabetic Retinopathy Database
  Calibration level 1}.
\newblock
\newblock


\bibitem[\protect\citeauthoryear{Kingma and Ba}{Kingma and Ba}{2014}]%
        {kingma2014adam}
\bibfield{author}{\bibinfo{person}{Diederik~P Kingma} {and}
  \bibinfo{person}{Jimmy Ba}.} \bibinfo{year}{2014}\natexlab{}.
\newblock \showarticletitle{Adam: A method for stochastic optimization}.
\newblock \bibinfo{journal}{\emph{arXiv preprint arXiv:1412.6980}}
  (\bibinfo{year}{2014}).
\newblock


\bibitem[\protect\citeauthoryear{Laserson, Lantsman, Cohen-Sfady, Tamir, Goz,
  Brestel, Bar, Atar, and Elnekave}{Laserson et~al\mbox{.}}{2018}]%
        {laserson2018textray}
\bibfield{author}{\bibinfo{person}{Jonathan Laserson},
  \bibinfo{person}{Christine~Dan Lantsman}, \bibinfo{person}{Michal
  Cohen-Sfady}, \bibinfo{person}{Itamar Tamir}, \bibinfo{person}{Eli Goz},
  \bibinfo{person}{Chen Brestel}, \bibinfo{person}{Shir Bar},
  \bibinfo{person}{Maya Atar}, {and} \bibinfo{person}{Eldad Elnekave}.}
  \bibinfo{year}{2018}\natexlab{}.
\newblock \showarticletitle{Textray: Mining clinical reports to gain a broad
  understanding of chest x-rays}. In \bibinfo{booktitle}{\emph{International
  Conference on Medical Image Computing and Computer-Assisted Intervention}}.
  Springer, \bibinfo{pages}{553--561}.
\newblock


\bibitem[\protect\citeauthoryear{Levy and Goldberg}{Levy and Goldberg}{2014a}]%
        {levy2014linguistic}
\bibfield{author}{\bibinfo{person}{Omer Levy} {and} \bibinfo{person}{Yoav
  Goldberg}.} \bibinfo{year}{2014}\natexlab{a}.
\newblock \showarticletitle{Linguistic regularities in sparse and explicit word
  representations}. In \bibinfo{booktitle}{\emph{Proceedings of the eighteenth
  conference on computational natural language learning}}.
  \bibinfo{pages}{171--180}.
\newblock


\bibitem[\protect\citeauthoryear{Levy and Goldberg}{Levy and Goldberg}{2014b}]%
        {levy2014neural}
\bibfield{author}{\bibinfo{person}{Omer Levy} {and} \bibinfo{person}{Yoav
  Goldberg}.} \bibinfo{year}{2014}\natexlab{b}.
\newblock \showarticletitle{Neural word embedding as implicit matrix
  factorization}.
\newblock \bibinfo{journal}{\emph{NIPS}}  \bibinfo{volume}{27}
  (\bibinfo{year}{2014}), \bibinfo{pages}{2177--2185}.
\newblock


\bibitem[\protect\citeauthoryear{Li, Liang, Hu, and Xing}{Li
  et~al\mbox{.}}{2018}]%
        {li2018hybrid}
\bibfield{author}{\bibinfo{person}{Yuan Li}, \bibinfo{person}{Xiaodan Liang},
  \bibinfo{person}{Zhiting Hu}, {and} \bibinfo{person}{Eric~P Xing}.}
  \bibinfo{year}{2018}\natexlab{}.
\newblock \showarticletitle{Hybrid retrieval-generation reinforced agent for
  medical image report generation}. In \bibinfo{booktitle}{\emph{Advances in
  Neural Information Processing Systems}}. \bibinfo{pages}{1530--1540}.
\newblock


\bibitem[\protect\citeauthoryear{Lin}{Lin}{2004}]%
        {lin2004rouge}
\bibfield{author}{\bibinfo{person}{Chin-Yew Lin}.}
  \bibinfo{year}{2004}\natexlab{}.
\newblock \showarticletitle{Rouge: A package for automatic evaluation of
  summaries}.
\newblock \bibinfo{journal}{\emph{Text Summarization Branches Out}}
  (\bibinfo{year}{2004}).
\newblock


\bibitem[\protect\citeauthoryear{Liu, Gardner, Belinkov, Peters, and Smith}{Liu
  et~al\mbox{.}}{2019}]%
        {liu2019linguistic}
\bibfield{author}{\bibinfo{person}{Nelson~F Liu}, \bibinfo{person}{Matt
  Gardner}, \bibinfo{person}{Yonatan Belinkov}, \bibinfo{person}{Matthew~E
  Peters}, {and} \bibinfo{person}{Noah~A Smith}.}
  \bibinfo{year}{2019}\natexlab{}.
\newblock \showarticletitle{Linguistic knowledge and transferability of
  contextual representations}.
\newblock \bibinfo{journal}{\emph{arXiv preprint arXiv:1903.08855}}
  (\bibinfo{year}{2019}).
\newblock


\bibitem[\protect\citeauthoryear{Liu, Zhu, Ye, Guadarrama, and Murphy}{Liu
  et~al\mbox{.}}{2017}]%
        {liu2017improved}
\bibfield{author}{\bibinfo{person}{Siqi Liu}, \bibinfo{person}{Zhenhai Zhu},
  \bibinfo{person}{Ning Ye}, \bibinfo{person}{Sergio Guadarrama}, {and}
  \bibinfo{person}{Kevin Murphy}.} \bibinfo{year}{2017}\natexlab{}.
\newblock \showarticletitle{Improved image captioning via policy gradient
  optimization of spider}. In \bibinfo{booktitle}{\emph{Proceedings of the IEEE
  international conference on computer vision}}. \bibinfo{pages}{873--881}.
\newblock


\bibitem[\protect\citeauthoryear{Liu, Yang, Yang, Huang, Tian, Morikawa, Tsai,
  and Tegner}{Liu et~al\mbox{.}}{2018}]%
        {liu2018synthesizing}
\bibfield{author}{\bibinfo{person}{Yi-Chieh Liu}, \bibinfo{person}{Hao-Hsiang
  Yang}, \bibinfo{person}{C-H~Huck Yang}, \bibinfo{person}{Jia-Hong Huang},
  \bibinfo{person}{Meng Tian}, \bibinfo{person}{Hiromasa Morikawa},
  \bibinfo{person}{Yi-Chang~James Tsai}, {and} \bibinfo{person}{Jesper
  Tegner}.} \bibinfo{year}{2018}\natexlab{}.
\newblock \showarticletitle{Synthesizing new retinal symptom images by multiple
  generative models}. In \bibinfo{booktitle}{\emph{Asian Conference on Computer
  Vision}}. Springer, \bibinfo{pages}{235--250}.
\newblock


\bibitem[\protect\citeauthoryear{Malinowski and Fritz}{Malinowski and
  Fritz}{2014}]%
        {malinowski2014multi}
\bibfield{author}{\bibinfo{person}{Mateusz Malinowski} {and}
  \bibinfo{person}{Mario Fritz}.} \bibinfo{year}{2014}\natexlab{}.
\newblock \showarticletitle{A multi-world approach to question answering about
  real-world scenes based on uncertain input}.
\newblock \bibinfo{journal}{\emph{arXiv preprint arXiv:1410.0210}}
  (\bibinfo{year}{2014}).
\newblock


\bibitem[\protect\citeauthoryear{Malinowski, Rohrbach, and Fritz}{Malinowski
  et~al\mbox{.}}{2015}]%
        {malinowski2015ask}
\bibfield{author}{\bibinfo{person}{Mateusz Malinowski}, \bibinfo{person}{Marcus
  Rohrbach}, {and} \bibinfo{person}{Mario Fritz}.}
  \bibinfo{year}{2015}\natexlab{}.
\newblock \showarticletitle{Ask your neurons: A neural-based approach to
  answering questions about images}. In \bibinfo{booktitle}{\emph{Proceedings
  of the IEEE international conference on computer vision}}.
  \bibinfo{pages}{1--9}.
\newblock


\bibitem[\protect\citeauthoryear{Malinowski, Rohrbach, and Fritz}{Malinowski
  et~al\mbox{.}}{2017}]%
        {malinowski2017ask}
\bibfield{author}{\bibinfo{person}{Mateusz Malinowski}, \bibinfo{person}{Marcus
  Rohrbach}, {and} \bibinfo{person}{Mario Fritz}.}
  \bibinfo{year}{2017}\natexlab{}.
\newblock \showarticletitle{Ask your neurons: A deep learning approach to
  visual question answering}.
\newblock \bibinfo{journal}{\emph{International Journal of Computer Vision}}
  \bibinfo{volume}{125}, \bibinfo{number}{1} (\bibinfo{year}{2017}),
  \bibinfo{pages}{110--135}.
\newblock


\bibitem[\protect\citeauthoryear{Mikolov, Sutskever, Chen, Corrado, and
  Dean}{Mikolov et~al\mbox{.}}{2013}]%
        {mikolov2013distributed}
\bibfield{author}{\bibinfo{person}{Tomas Mikolov}, \bibinfo{person}{Ilya
  Sutskever}, \bibinfo{person}{Kai Chen}, \bibinfo{person}{Greg Corrado}, {and}
  \bibinfo{person}{Jeffrey Dean}.} \bibinfo{year}{2013}\natexlab{}.
\newblock \showarticletitle{Distributed representations of words and phrases
  and their compositionality}.
\newblock \bibinfo{journal}{\emph{arXiv preprint arXiv:1310.4546}}
  (\bibinfo{year}{2013}).
\newblock


\bibitem[\protect\citeauthoryear{Niemeijer, Xu, Dumitrescu, Gupta, van
  Ginneken, Folk, and Abramoff}{Niemeijer et~al\mbox{.}}{2011}]%
        {niemeijer2011inspire}
\bibfield{author}{\bibinfo{person}{M Niemeijer}, \bibinfo{person}{X Xu},
  \bibinfo{person}{A Dumitrescu}, \bibinfo{person}{P Gupta}, \bibinfo{person}{B
  van Ginneken}, \bibinfo{person}{J Folk}, {and} \bibinfo{person}{M Abramoff}.}
  \bibinfo{year}{2011}\natexlab{}.
\newblock \bibinfo{title}{INSPIRE-AVR: Iowa normative set for processing images
  of the retina-artery vein ratio}.
\newblock
\newblock


\bibitem[\protect\citeauthoryear{Ortega, Penedo, Rouco, Barreira, and
  Carreira}{Ortega et~al\mbox{.}}{2009}]%
        {ortega2009retinal}
\bibfield{author}{\bibinfo{person}{Marcos Ortega}, \bibinfo{person}{Manuel~G
  Penedo}, \bibinfo{person}{Jos{\'e} Rouco}, \bibinfo{person}{Noelia Barreira},
  {and} \bibinfo{person}{Mar{\'\i}a~J Carreira}.}
  \bibinfo{year}{2009}\natexlab{}.
\newblock \showarticletitle{Retinal verification using a feature points-based
  biometric pattern}.
\newblock \bibinfo{journal}{\emph{EURASIP Journal on Advances in Signal
  Processing}}  \bibinfo{volume}{2009} (\bibinfo{year}{2009}),
  \bibinfo{pages}{2}.
\newblock


\bibitem[\protect\citeauthoryear{Pan, Yao, Li, and Mei}{Pan
  et~al\mbox{.}}{2020}]%
        {pan2020x}
\bibfield{author}{\bibinfo{person}{Yingwei Pan}, \bibinfo{person}{Ting Yao},
  \bibinfo{person}{Yehao Li}, {and} \bibinfo{person}{Tao Mei}.}
  \bibinfo{year}{2020}\natexlab{}.
\newblock \showarticletitle{X-Linear Attention Networks for Image Captioning}.
  In \bibinfo{booktitle}{\emph{Proceedings of the IEEE/CVF Conference on
  Computer Vision and Pattern Recognition}}. \bibinfo{pages}{10971--10980}.
\newblock


\bibitem[\protect\citeauthoryear{Papineni, Roukos, Ward, and Zhu}{Papineni
  et~al\mbox{.}}{2002}]%
        {papineni2002bleu}
\bibfield{author}{\bibinfo{person}{Kishore Papineni}, \bibinfo{person}{Salim
  Roukos}, \bibinfo{person}{Todd Ward}, {and} \bibinfo{person}{Wei-Jing Zhu}.}
  \bibinfo{year}{2002}\natexlab{}.
\newblock \showarticletitle{BLEU: a method for automatic evaluation of machine
  translation}. In \bibinfo{booktitle}{\emph{Proceedings of ACL}}. Association
  for Computational Linguistics, \bibinfo{pages}{311--318}.
\newblock


\bibitem[\protect\citeauthoryear{Pennington, Socher, and Manning}{Pennington
  et~al\mbox{.}}{2014}]%
        {pennington2014glove}
\bibfield{author}{\bibinfo{person}{Jeffrey Pennington},
  \bibinfo{person}{Richard Socher}, {and} \bibinfo{person}{Christopher~D
  Manning}.} \bibinfo{year}{2014}\natexlab{}.
\newblock \showarticletitle{Glove: Global vectors for word representation}. In
  \bibinfo{booktitle}{\emph{Proceedings of the 2014 conference on empirical
  methods in natural language processing (EMNLP)}}.
  \bibinfo{pages}{1532--1543}.
\newblock


\bibitem[\protect\citeauthoryear{Peters, Neumann, Iyyer, Gardner, Clark, Lee,
  and Zettlemoyer}{Peters et~al\mbox{.}}{2018}]%
        {peters2018deep}
\bibfield{author}{\bibinfo{person}{Matthew~E Peters}, \bibinfo{person}{Mark
  Neumann}, \bibinfo{person}{Mohit Iyyer}, \bibinfo{person}{Matt Gardner},
  \bibinfo{person}{Christopher Clark}, \bibinfo{person}{Kenton Lee}, {and}
  \bibinfo{person}{Luke Zettlemoyer}.} \bibinfo{year}{2018}\natexlab{}.
\newblock \showarticletitle{Deep contextualized word representations}.
\newblock \bibinfo{journal}{\emph{arXiv preprint arXiv:1802.05365}}
  (\bibinfo{year}{2018}).
\newblock


\bibitem[\protect\citeauthoryear{Porwal, Pachade, Kamble, Kokare, Deshmukh,
  Sahasrabuddhe, and Meriaudeau}{Porwal et~al\mbox{.}}{2018}]%
        {porwal2018indian}
\bibfield{author}{\bibinfo{person}{Prasanna Porwal}, \bibinfo{person}{Samiksha
  Pachade}, \bibinfo{person}{Ravi Kamble}, \bibinfo{person}{Manesh Kokare},
  \bibinfo{person}{Girish Deshmukh}, \bibinfo{person}{Vivek Sahasrabuddhe},
  {and} \bibinfo{person}{Fabrice Meriaudeau}.} \bibinfo{year}{2018}\natexlab{}.
\newblock \showarticletitle{Indian diabetic retinopathy image dataset (IDRiD):
  a database for diabetic retinopathy screening research}.
\newblock \bibinfo{journal}{\emph{Data}} \bibinfo{volume}{3},
  \bibinfo{number}{3} (\bibinfo{year}{2018}), \bibinfo{pages}{25}.
\newblock


\bibitem[\protect\citeauthoryear{Radford, Wu, Child, Luan, Amodei, and
  Sutskever}{Radford et~al\mbox{.}}{2019}]%
        {radford2019language}
\bibfield{author}{\bibinfo{person}{Alec Radford}, \bibinfo{person}{Jeffrey Wu},
  \bibinfo{person}{Rewon Child}, \bibinfo{person}{David Luan},
  \bibinfo{person}{Dario Amodei}, {and} \bibinfo{person}{Ilya Sutskever}.}
  \bibinfo{year}{2019}\natexlab{}.
\newblock \showarticletitle{Language models are unsupervised multitask
  learners}.
\newblock \bibinfo{journal}{\emph{OpenAI blog}} \bibinfo{volume}{1},
  \bibinfo{number}{8} (\bibinfo{year}{2019}), \bibinfo{pages}{9}.
\newblock


\bibitem[\protect\citeauthoryear{Russakovsky, Deng, Su, Krause, Satheesh, Ma,
  Huang, Karpathy, Khosla, Bernstein, et~al\mbox{.}}{Russakovsky
  et~al\mbox{.}}{2015}]%
        {russakovsky2015imagenet}
\bibfield{author}{\bibinfo{person}{Olga Russakovsky}, \bibinfo{person}{Jia
  Deng}, \bibinfo{person}{Hao Su}, \bibinfo{person}{Jonathan Krause},
  \bibinfo{person}{Sanjeev Satheesh}, \bibinfo{person}{Sean Ma},
  \bibinfo{person}{Zhiheng Huang}, \bibinfo{person}{Andrej Karpathy},
  \bibinfo{person}{Aditya Khosla}, \bibinfo{person}{Michael Bernstein},
  {et~al\mbox{.}}} \bibinfo{year}{2015}\natexlab{}.
\newblock \showarticletitle{Imagenet large scale visual recognition challenge}.
\newblock \bibinfo{journal}{\emph{International journal of computer vision}}
  \bibinfo{volume}{115}, \bibinfo{number}{3} (\bibinfo{year}{2015}),
  \bibinfo{pages}{211--252}.
\newblock


\bibitem[\protect\citeauthoryear{Scott and Matwin}{Scott and Matwin}{1998}]%
        {scott1998text}
\bibfield{author}{\bibinfo{person}{Sam Scott} {and} \bibinfo{person}{Stan
  Matwin}.} \bibinfo{year}{1998}\natexlab{}.
\newblock \showarticletitle{Text classification using WordNet hypernyms}. In
  \bibinfo{booktitle}{\emph{Usage of WordNet in Natural Language Processing
  Systems}}.
\newblock


\bibitem[\protect\citeauthoryear{Simonyan and Zisserman}{Simonyan and
  Zisserman}{2014}]%
        {simonyan2014very}
\bibfield{author}{\bibinfo{person}{Karen Simonyan} {and}
  \bibinfo{person}{Andrew Zisserman}.} \bibinfo{year}{2014}\natexlab{}.
\newblock \showarticletitle{Very deep convolutional networks for large-scale
  image recognition}.
\newblock \bibinfo{journal}{\emph{arXiv preprint arXiv:1409.1556}}
  (\bibinfo{year}{2014}).
\newblock


\bibitem[\protect\citeauthoryear{Sivaswamy, Krishnadas, Joshi, Jain, and
  Tabish}{Sivaswamy et~al\mbox{.}}{2014}]%
        {sivaswamy2014drishti}
\bibfield{author}{\bibinfo{person}{Jayanthi Sivaswamy}, \bibinfo{person}{SR
  Krishnadas}, \bibinfo{person}{Gopal~Datt Joshi}, \bibinfo{person}{Madhulika
  Jain}, {and} \bibinfo{person}{A~Ujjwaft~Syed Tabish}.}
  \bibinfo{year}{2014}\natexlab{}.
\newblock \showarticletitle{Drishti-gs: Retinal image dataset for optic nerve
  head (onh) segmentation}. In \bibinfo{booktitle}{\emph{2014 IEEE 11th
  International Symposium on Biomedical Imaging (ISBI)}}. IEEE,
  \bibinfo{pages}{53--56}.
\newblock


\bibitem[\protect\citeauthoryear{Soumya~George and Joseph}{Soumya~George and
  Joseph}{2014}]%
        {soumya2014text}
\bibfield{author}{\bibinfo{person}{K Soumya~George} {and}
  \bibinfo{person}{Shibily Joseph}.} \bibinfo{year}{2014}\natexlab{}.
\newblock \showarticletitle{Text classification by augmenting bag of words
  (BOW) representation with co-occurrence feature}.
\newblock \bibinfo{journal}{\emph{IOSR J. Comput. Eng}} \bibinfo{volume}{16},
  \bibinfo{number}{1} (\bibinfo{year}{2014}), \bibinfo{pages}{34--38}.
\newblock


\bibitem[\protect\citeauthoryear{Staal, Abr{\`a}moff, Niemeijer, Viergever, and
  Van~Ginneken}{Staal et~al\mbox{.}}{2004}]%
        {staal2004ridge}
\bibfield{author}{\bibinfo{person}{Joes Staal}, \bibinfo{person}{Michael~D
  Abr{\`a}moff}, \bibinfo{person}{Meindert Niemeijer}, \bibinfo{person}{Max~A
  Viergever}, {and} \bibinfo{person}{Bram Van~Ginneken}.}
  \bibinfo{year}{2004}\natexlab{}.
\newblock \showarticletitle{Ridge-based vessel segmentation in color images of
  the retina}.
\newblock \bibinfo{journal}{\emph{TMI}} \bibinfo{volume}{23},
  \bibinfo{number}{4} (\bibinfo{year}{2004}), \bibinfo{pages}{501--509}.
\newblock


\bibitem[\protect\citeauthoryear{Vaswani, Shazeer, Parmar, Uszkoreit, Jones,
  Gomez, Kaiser, and Polosukhin}{Vaswani et~al\mbox{.}}{2017}]%
        {vaswani2017attention}
\bibfield{author}{\bibinfo{person}{Ashish Vaswani}, \bibinfo{person}{Noam
  Shazeer}, \bibinfo{person}{Niki Parmar}, \bibinfo{person}{Jakob Uszkoreit},
  \bibinfo{person}{Llion Jones}, \bibinfo{person}{Aidan~N Gomez},
  \bibinfo{person}{Lukasz Kaiser}, {and} \bibinfo{person}{Illia Polosukhin}.}
  \bibinfo{year}{2017}\natexlab{}.
\newblock \showarticletitle{Attention is all you need}.
\newblock \bibinfo{journal}{\emph{arXiv preprint arXiv:1706.03762}}
  (\bibinfo{year}{2017}).
\newblock


\bibitem[\protect\citeauthoryear{V{\'a}zquez, Cancela, Barreira, Penedo,
  Rodr{\'\i}guez-Blanco, Seijo, de~Tuero, Barcel{\'o}, and Saez}{V{\'a}zquez
  et~al\mbox{.}}{2013}]%
        {vazquez2013improving}
\bibfield{author}{\bibinfo{person}{SG V{\'a}zquez}, \bibinfo{person}{Brais
  Cancela}, \bibinfo{person}{Noelia Barreira}, \bibinfo{person}{Manuel~G
  Penedo}, \bibinfo{person}{M Rodr{\'\i}guez-Blanco}, \bibinfo{person}{M~Pena
  Seijo}, \bibinfo{person}{G~Coll de Tuero}, \bibinfo{person}{Maria~Ant{\`o}nia
  Barcel{\'o}}, {and} \bibinfo{person}{Marc Saez}.}
  \bibinfo{year}{2013}\natexlab{}.
\newblock \showarticletitle{Improving retinal artery and vein classification by
  means of a minimal path approach}.
\newblock \bibinfo{journal}{\emph{Machine vision and applications}}
  \bibinfo{volume}{24}, \bibinfo{number}{5} (\bibinfo{year}{2013}),
  \bibinfo{pages}{919--930}.
\newblock


\bibitem[\protect\citeauthoryear{Vedantam, Lawrence~Zitnick, and
  Parikh}{Vedantam et~al\mbox{.}}{2015}]%
        {vedantam2015cider}
\bibfield{author}{\bibinfo{person}{Ramakrishna Vedantam}, \bibinfo{person}{C
  Lawrence~Zitnick}, {and} \bibinfo{person}{Devi Parikh}.}
  \bibinfo{year}{2015}\natexlab{}.
\newblock \showarticletitle{Cider: Consensus-based image description
  evaluation}. In \bibinfo{booktitle}{\emph{Proceedings of the IEEE conference
  on computer vision and pattern recognition}}. \bibinfo{pages}{4566--4575}.
\newblock


\bibitem[\protect\citeauthoryear{Vinyals, Toshev, Bengio, and Erhan}{Vinyals
  et~al\mbox{.}}{2015}]%
        {vinyals2015show}
\bibfield{author}{\bibinfo{person}{Oriol Vinyals}, \bibinfo{person}{Alexander
  Toshev}, \bibinfo{person}{Samy Bengio}, {and} \bibinfo{person}{Dumitru
  Erhan}.} \bibinfo{year}{2015}\natexlab{}.
\newblock \showarticletitle{Show and tell: A neural image caption generator}.
  In \bibinfo{booktitle}{\emph{2015 IEEE Computer Society Conference on
  Computer Vision and Pattern Recognition (CVPR'15)}}.
  \bibinfo{pages}{3156--3164}.
\newblock


\bibitem[\protect\citeauthoryear{Yang, Huang, Liu, Chiu, Gao, Lyu, Tegner,
  et~al\mbox{.}}{Yang et~al\mbox{.}}{2018a}]%
        {yang2018novel}
\bibfield{author}{\bibinfo{person}{C-H~Huck Yang}, \bibinfo{person}{Jia-Hong
  Huang}, \bibinfo{person}{Fangyu Liu}, \bibinfo{person}{Fang-Yi Chiu},
  \bibinfo{person}{Mengya Gao}, \bibinfo{person}{Weifeng Lyu},
  \bibinfo{person}{Jesper Tegner}, {et~al\mbox{.}}}
  \bibinfo{year}{2018}\natexlab{a}.
\newblock \showarticletitle{A novel hybrid machine learning model for
  auto-classification of retinal diseases}.
\newblock \bibinfo{journal}{\emph{ICML Workshop on Computational Biology}}
  (\bibinfo{year}{2018}).
\newblock


\bibitem[\protect\citeauthoryear{Yang, Liu, Huang, Tian, Lin, Liu, Morikawa,
  Yang, and Tegner}{Yang et~al\mbox{.}}{2018b}]%
        {yang2018auto}
\bibfield{author}{\bibinfo{person}{C-H~Huck Yang}, \bibinfo{person}{Fangyu
  Liu}, \bibinfo{person}{Jia-Hong Huang}, \bibinfo{person}{Meng Tian},
  \bibinfo{person}{MD~I-Hung Lin}, \bibinfo{person}{Yi~Chieh Liu},
  \bibinfo{person}{Hiromasa Morikawa}, \bibinfo{person}{Hao-Hsiang Yang}, {and}
  \bibinfo{person}{Jesper Tegner}.} \bibinfo{year}{2018}\natexlab{b}.
\newblock \showarticletitle{Auto-classification of retinal diseases in the
  limit of sparse data using a two-streams machine learning model}. In
  \bibinfo{booktitle}{\emph{Asian Conference on Computer Vision}}. Springer,
  \bibinfo{pages}{323--338}.
\newblock


\end{thebibliography}

%
\appendix

\end{document}